%% file: paper.tex
\newcommand{\PaperTitle}{Are Open-Weight LLMs Ready for Social Media Moderation? A Comparative Study on Bluesky}
\newcommand{\PaperNumber}{143}

\documentclass[10pt,sigconf,nonacm]{acmart}

\usepackage[]{dirtytalk}  
\usepackage{csquotes}     
\usepackage{siunitx}      
\usepackage{subcaption}   
\usepackage{enumitem}     

\newcommand{\CrawlerName}{Ouranos}
\newcommand{\CrawlerNaming}{Ouranos(,
    also Uranus,)
    is the personification of the sky and one of the Greek primordial deities;
    it is also the name of the blue planet, the seventh from the Sun.}

\author{Hsuan-Yu \textsc{Chou}}
\orcid{0009-0005-6894-9364}
\affiliation{
  \institution{Duke University}
  \country{USA}
}
\email{hsuan-yu.chou@duke.edu}

\author{Wajiha Naveed}
\affiliation{
  \institution{Duke University}
  \country{USA}
}
\email{wajiha.naveed@duke.edu}

\author{Shuyan Zhou}
\affiliation{
  \institution{Duke University}
  \country{USA}
}
\email{shuyan.zhou@duke.edu}

\author{Xiaowei Yang}
\affiliation{
  \institution{Duke University}
  \country{USA}
}
\email{xwy@cs.duke.edu}

\begin{document}

\title{\PaperTitle}

\begin{abstract}
  As internet access expands,
  so does exposure to harmful content,
  increasing the need for effective moderation.
  Research has demonstrated that
  large language models (LLMs) can be effectively utilized for social media moderation tasks,
  including harmful content detection.
  While proprietary LLMs have been shown to zero-shot outperform traditional machine learning models,
  the out-of-the-box capability of open-weight LLMs remains an open question.

  Motivated by recent developments of reasoning LLMs,
  we evaluate seven state-of-the-art models:
  four proprietary and three open-weight.
  Testing with real-world posts on Bluesky,
  moderation decisions by Bluesky Moderation Service,
  and annotations by two authors,
  we find a considerable degree of overlap between
  the sensitivity (81\%--97\%) and specificity (91\%--100\%) of the open-weight LLMs
  and those (72\%--98\%, and 93\%--99\%) of the proprietary ones.
  Additionally,
  our analysis reveals that
  specificity exceeds sensitivity for rudeness detection,
  but the opposite holds for intolerance and threats.
  Lastly,
  we identify inter-rater agreement across human moderators and the LLMs,
  highlighting considerations for deploying LLMs
  in both platform-scale and personalized moderation contexts.
  These findings show open-weight LLMs can support privacy-preserving moderation on consumer-grade hardware
  and suggest new directions for designing moderation systems
  that balance community values with individual user preferences.
\end{abstract}

\settopmatter{printfolios=true}  
\maketitle

\input{sec_introduction}
\input{sec_related}
\input{sec_methods}
\input{sec_results}
\input{sec_discussion}
\input{sec_acknowledgements}


\appendix

\input{sec_ethics}

\input{sec_misc}


\bibliographystyle{ACM-Reference-Format}
\bibliography{bibs}

\end{document}

%% file: sec_introduction.tex
\section{Introduction}

Due to the ubiquity of harmful content on the Internet,
content moderation is key to online safety~\cite{
    bhargava_gobo_2019,
    guo_moderating_2024,
    la_cava_safeguarding_2025,
    lees_new_2022,
    papadamou_disturbed_2020,
    vishwamitra_moderating_2024,
    yousaf_deep_2022}.
Automated content moderation using machine learning (ML) techniques
has been previously studied~\cite{
    guo_investigation_2023,
    lees_new_2022,
    papadamou_disturbed_2020,
    yousaf_deep_2022,
    zhan_slm-mod_2025}.
The release of ChatGPT by OpenAI
led on to the rapid growth of
the popularity of large language models (LLMs) among the general public
and the interest of potential applications within the research community.
Multiple studies have reported findings of using LLMs
to perform common social media content moderation tasks,
such as detecting harmful content~\cite{
    guo_investigation_2023,
    guo_moderating_2024,
    kumar_watch_2024,
    vishwamitra_moderating_2024,
    zhan_slm-mod_2025}
and enforcing community rules~\cite{
    kolla_llm-mod_2024,
    kumar_watch_2024,
    la_cava_safeguarding_2025}.
Although past evaluations yielded mixed results in the case of community rule compliance checks,
they were unequivocally positive for harmful content detection:
Proprietary LLMs surpassed existing state-of-the-art commercial tools in terms of accuracy~\cite{
    guo_moderating_2024,
    kumar_watch_2024,
    vishwamitra_moderating_2024}.

In addition to the rapid development of proprietary LLMs,
open-weight LLMs
(LLMs whose model weights are open,
hence can be fine-tuned and deployed locally at users' will),
have also gained traction~\cite{
    deepseek-ai_deepseek-r1_2025}.
Locally-hosted LLMs are suitable in areas where there are privacy concerns~\cite{
    hanke_open_2024,
    huang_-premises_2025}
(e.g. medical applications~\cite{
    nowak_privacy-ensuring_2025}).
However,
prior work had reported that,
unlike proprietary LLMs,
open-weight LLMs that could run on consumer-grade devices
either were inadequate for harmful content detection~\cite{
    kumar_watch_2024},
or required fine-tuning to achieve satisfiable accuracy~\cite{
    zhan_slm-mod_2025}.
Recently,
latest open-weight models have been released with reasoning ability~\cite{
    gibney_openai_2025,
    chen_towards_2025},
but their applicability for content moderation has not been studied.
So we asked this research question:
\textbf{Are open-weight LLMs now as good as proprietary ones out-of-the-box
    at detecting harmful content on social media?}

To answer the question,
we started by building a ground-truth dataset.
We collected real-world posts from Bluesky~\cite{
    kleppmann_bluesky_2024}
and moderation decisions made by the Bluesky Moderation Service (BMS).
We sampled
three categories
(namely \emph{rude}, \emph{intolerant}, and \emph{threat})
of rule-violating posts
and randomly sampled posts not flagged by BMS.
For each of the posts sampled,
two authors independently annotated whether or not it meets the moderation
criteria set by BMS~\cite{
    bluesky_moderation_service_bluesky_nodate}.
(More in Section~\ref{sec:data}.)

Next,
for evaluation,
we selected seven state-of-the-art LLMs.
Three of them
are open-weight reasoning LLMs
that can each fit into a single consumer-grade GPU card with 24-GB VRAM;
three of them
are proprietary reasoning LLMs;
and we pick one non-reasoning proprietary LLM
as a contrast.
Applying the same moderation criteria given by BMS,
we prompted the LLMs to perform content moderation,
i.e. annotate the posts the same way as the two author annotators did.
(More in Section~\ref{sec:model}.)

Finally,
we processed the model outputs and analyzed the data.
We observe that
treating the annotations made by human
(BMS and the two author annotators)
as ground truth,
open-weight LLMs and proprietary LLMs exhibit overlapping accuracy
(81\%--97\% and 72\%--98\% for sensitivity,
and 91\%--100\% and 93\%--99\% for specificity).
Interestingly,
generally speaking,
the \emph{rude} test yields a higher specificity than sensitivity,
while the opposite is true for the \emph{intolerant} and \emph{threat} tests.
(More in Section~\ref{sec:results}.)

Our results show that
open-weight reasoning LLMs can detect harmful content on social media
as accurately as proprietary models,
even without fine-tuning.
We discuss the practical implications, limitations, and directions for future work in Section~\ref{sec:discussion}.

\subsection*{Key contributions}

The key contributions of our work are as follows:

\begin{enumerate}[leftmargin=*]
    \item \textbf{New labeled dataset of real-world social media content moderation.}
          We contributed a new labeled dataset of real-world social media posts on Bluesky combining official moderation decisions by Bluesky Moderation Service with independent author annotations.

    \item \textbf{New understanding of the out-of-the-box competence of open-weight LLMs at zero-shot harmful content detection.}
          We show that
          without the need to fine-tune,
          state-of-the-art open-weight reasoning LLMs
          (that can fit in the VRAM of a single consumer-grade GPU card)
          have comparable accuracy
          (both sensitivity and specificity)
          to larger proprietary LLMs
          in terms of zero-shot text-based harmful content detection,
          overturning the findings of previous studies~\cite{
            kumar_watch_2024,
            zhan_slm-mod_2025}.
    \item \textbf{Attribute-specific performance patterns that motivate standardization.}
        Accuracy is not uniform across harm categories.
        On posts unanimously labeled by humans,
        rudeness detection shows higher sensitivity than specificity (about 95\%--98\% vs. 91\%--96\%),
        whereas intolerance and threats show near-perfect specificity (about 98\%--100\%)
        but markedly lower sensitivity.
        These distinct error profiles arise from differences in category definitions and review criteria.
        This finding underscores the need for attribute-specific taxonomies,
        illustrative examples,
        and calibrated decision thresholds
        to reduce label ambiguity
        and align moderation behavior with community norms and user preferences.
\end{enumerate}

%% file: sec_related.tex
\section{Background and Related Work}

Social media content moderation has been an active area of research.
From traditional machine learning (ML) classifiers to large language models (LLMs),
many studies have investigated automated content moderation systems.
In this section,
we provide the necessary context
and summarize prior studies that inspire our research question.


Perspective API~\cite{lees_new_2022,google_jigsaw_perspective_nodate}
is a moderation service developed and provided by Google Jigsaw.
Widely adopted by online platforms and extensively utilized by the research community~\cite{
    kumar_watch_2024,
    nogara_toxic_2025},
Perspective API's ML model takes text as input,
and produces scores between 0 and 1 for attributes
such as toxicity, identity attack, threat, etc.


\citeauthor{guo_investigation_2023}~\cite{guo_investigation_2023}
showed that
LLMs (GPT-3.5-turbo) can match and often outperform ML models
such as BERT and RoBERTa
in detecting hate speech,
leveraging five existing benchmarks of diverse hate speech categories.
\citeauthor{vishwamitra_moderating_2024}~\cite{vishwamitra_moderating_2024}
introduced HateGuard,
a framework
that employs a custom chain-of-thought~\cite{
    wei_chain--thought_2022}
prompting strategy to utilize LLMs for newfangled hate speech detection.
In comparison to existing methods employing ML models,
HateGuard makes adaptation to new hate terms faster and has higher F1 scores
(90\%--99\% using GPT-4)
in detecting various new forms of hate speech on Twitter.

\citeauthor{kumar_watch_2024}~\cite{kumar_watch_2024}
systematically investigated content moderation using LLMs.
The authors evaluated five LLMs
(GPT-3, GPT-3.5, GPT-4, Gemini Pro, and LLAMA 2)
on rule-based moderation and toxicity detection.
For rule-based moderation,
they collected 95 subreddit community rules and for each subreddit,
500 comments removed by moderators (sourced from a prior study)
and 500 not removed (queried from Reddit API);
whereas for toxic content detection,
they sampled 5000 toxic comments and 5000 non toxic ones
from a dataset with comments collected from Reddit, Twitter, and 4chan
and annotated in a prior study.
They found that proprietary LLMs
achieve comparable performance to human moderators in rule-based moderation tasks
and outperform Perspective API significantly in toxic content detection tasks
(with accuracy spanning from 70\% to 73\% and sensitivity 73\% to 86\%).
However,
they reported that
(with an accuracy of 63\% and a precision of 59\%,)
LLAMA 2,
the open-weight LLM,
is incapable of performing toxic content detection.

Around the same period,
\citeauthor{kolla_llm-mod_2024}~\cite{kolla_llm-mod_2024}
also explored using LLMs to identify community rule violations on Reddit.
The authors created a dataset with 144 rule-violating posts
(36 manually selected from Reddit,
47 written by the authors,
and 61 generated by GPT-3.5)
and 600
(popular hence assumed to be)
rule-passing posts
(collected from Reddit API).
However,
they reported that
while the GPT-3.5 has a high specificity (92.3\%),
it has a low sensitivity (43.1\%).

In a similar theme,
\citeauthor{la_cava_safeguarding_2025}~\cite{la_cava_safeguarding_2025}
looked into automated community rule compliance checking for decentralized social networks.
The authors collected community rules from 508 Mastodon~\cite{
    mastodon_ggmbh_mastodon_nodate}
servers,
and for each server,
collected 100 posts using the API of Mastodon.
They utilized six open-weight LLMs
(gemma-2-9b-it,
Llama-3.1-8B-Instruct,
Mistral-7B-Instruct-v0.3,
Mistral-Nemo-Instruct-2407,
Neural-Chat-v3-3,
and Qwen2-7B-Instruct)
to grade community rule compliance
and generate justification and suggestion for improvement.
They reported positive results from qualitative and quantitative evaluations.

More recently,
\citeauthor{zhan_slm-mod_2025}~\cite{zhan_slm-mod_2025}
also explored the performance of open-weight LLMs
for community-specific content moderation tasks.
The authors collected 5000 removed and 5000 not-removed Reddit comments
(, from each of 15 subreddits),
sampled 8000 for finetuning and 2000 for evaluation.
Using the LoRA~\cite{
    hu_lora_2021}
technique,
they fine-tuned LLMs with fewer than 15B parameters
(Llama-3.1-8b,
Gemma-2-9b,
and Mistral-nemo-instruct),
and evaluated them against one larger open-weight LLM
(Cohere's Command R+)
and two proprietary LLMs
(GPT-4o and GPT-4o-mini).
They found that
the accuracy
(72.5\%--77.87\%)
of fine-tuned smaller LLMs consistently exceeds that of the larger LLMs
(by an average of 11.5\%).
They also reported that
larger LLMs have higher precision than recall,
and smaller LLMs the opposite.




%% file: sec_methods.tex
\section{Methodology}

\subsection{Data collection}\label{sec:data}

Prior studies often relied on Twitter (now X) as a major source of real-world data~\cite{
    guo_moderating_2024,
    kumar_watch_2024,
    vishwamitra_moderating_2024}.
After Twitter restricted public content access in 2023,
we sought an alternative data source.
We selected Bluesky~\cite{
    kleppmann_bluesky_2024},
a Twitter-like microblogging platform that remains publicly accessible and has a relatively active user base (\emph{cf.} Mastodon).
we decided to turn to Bluesky~\cite{kleppmann_bluesky_2024},
a microblogging social media platform that resembles Twitter,
allows public access,
and has a relatively large active user base (\emph{cf} Mastodon).

Bluesky,
perhaps due to its young age,
was designed with moderation in mind.
Bluesky's stackable approach to moderation\footnote{https://bsky.social/about/blog/03-12-2024-stackable-moderation} means that everyone can become a moderation service provider,
and that everyone can subscribe to their choice of moderation services.
For the research community,
this means that in addition to access to content (posts and comments) and interactions (follows, likes, and reposts) on Bluesky,
we also have access to (previous difficult to obtain) moderator decisions.

\subsubsection{Posts on Bluesky}

Unlike traditional centralized social media platforms,
where user-generated content are managed and distributed by the platforms,
Bluesky users keep their data
(including posts, likes, follows, etc.)
on \emph{Personal Data Servers (PDS)} of their choosing
(or host their own PDSes),
and Bluesky's indexing infrastructure
(i.e. Bluesky \emph{Relay} and \emph{App View})
subscribes to known PDSes
and distributes the content generated by users.
This AT Protocol design implies that
anyone can access all posts authored by known users via the authors' PDSes.

We obtained a list of known Bluesky accounts
utilizing Skyfall~\cite{mccain_stanfordioskyfall_2025},
an open-source toolkit written in Go,
which sends \texttt{\footnotesize com.atproto.sync.listRepos} queries
to the Bluesky \emph{Relay} at \url{https://bsky.network}.
To obtain a list of known Bluesky posts
we referenced Skyfall and developed a specialized toolkit called \CrawlerName{}\footnote{
    \CrawlerNaming{}
    We plan to open source \CrawlerName{} after the paper review process.
}.
For each unique account,
\CrawlerName{} requests all posts authored by the account
by sending \texttt{\footnotesize app.bsky.feed.getAuthorFeed} queries
to Bluesky's index server at \url{https://public.api.bsky.app}.
Note that
to minimize the impact of our data collection process,
instead of querying each user's PDS server,
we chose to query Bluesky's index server,
which employs content delivery network (CDN) for caching.

We took a snapshot of posts on Bluesky from the 12th to the 24th day of August, 2025.
We discovered about \num{1.6e9} posts
from around \num{2.5e7} unique accounts\footnote{
    At the time of collection,
    Bluesky operated three \emph{Relays}.
    We discovered \num{25321000} accounts from \url{https://bsky.network},
    \num{20000} from \url{https://relay1.us-east.bsky.network},
    and \num{35000} from \url{https://relay1.us-west.bsky.network}.
    The latter two \emph{Relays} were introduced in May, 2025
    as part of Bluesky's test for protocol update~\cite{bluesky_pbc_relay_2025}.
    The lists from the latter two \emph{Relays} are subsets of the first.
}.
Due to limited resources,
We randomly sampled 4\% of accounts (\num{1012840} accounts),
which authored \num{6.6e7} posts,
for analysis.
Since we focus on text-based content moderation in this study,
we filtered out posts with multimedia content.
To obtain a more consistent distribution within the confined dataset,
we limited the scope of our dataset to root-level posts(,
i.e. excluding replies to other posts),
and filtered out posts that include mentions or hyperlinks,
and posts that are not exclusively in English.
The remaining are \num{4339221} text-only root posts in English,
which are from \num{141506} unique accounts.

\subsubsection{Labels by Bluesky Moderation Service}

In addition to posts,
we collected public moderation decisions
(\emph{labels})
published by Bluesky Moderation Service (BMS),
the official moderation service operated by Bluesky.
By subscribing to BMS's labeler endpoint at \url{wss://mod.bsky.app/xrpc}.
we collected more than \num{22e6} labels,
which were published between April, 2023 and October, 2025.

We selected three labels most-applicable to our study:
\emph{rude}, \emph{intolerant}, and \emph{threat}.
Although BMS issues 67 unique types of labels\footnote{
    The most frequently applied are \emph{porn} and \emph{sexual}.},
we chose the most commonly seen label values in our sampled dataset.
For the \num{4339221} text-only root posts in English,
the top-5 are
\emph{intolerant} (202),
\emph{!hide} (29),
\emph{threat} (33),
\emph{rude} (26),
and \emph{!warn} (5).
We excluded label values \emph{!hide} and \emph{!warn} from our study
because BMS does not provide descriptions for them\footnote{
    Label values start with an exclamation point (\emph{!}) are reserved by Bluesky.
    The label values \emph{!hide} and \emph{!warn} are two of the seven values defined in AT Protocol.
    They trigger specific UI actions in the client application developed by Bluesky.
    See~\cite{bluesky_pbc_labels_nodate} for details.
}.
We then utilized \CrawlerName{} to expand our sampled dataset
by sampling posts with the labels of interest from the full dataset
until we have at least 520 posts for each label.

One challenge we faced is
whether the BMS labels were AI-generated
or ground-truth labels created by human moderators\footnote{
    Bluesky has mentioned in a blog post~\cite{
        the_bluesky_team_share_2024}
    that \say{[e]very video is processed via Hive and Thorn
        to scan for content that requires a content warning
        or content that should be taken down
        (e.g. illegal material like CSAM).}
}.
To answer this question,
we calculated the Labeler Reaction Latency~\cite{balduf_looking_2024}.
Figure \ref{fig:ecdf-labeler-latency_rude} shows the labeler reaction latency,
i.e. time difference between when a post was created\footnote{
    We take the earlier of the \texttt{createdAt} timestamp
    (set by user)
    and the \texttt{indexedAt} timestamp
    (set by Bluesky's indexing infrastructure).
} and when BMS labeled the post.
More than 80\% of posts were labeled at least one day after the posts were created.
Therefore,
we believe the labels are authentic human moderation decisions,
since it is unlikely that the entire labeling pipeline is automated
while the latency is long and inconsistent.

\input{fig_ecdf-labeler-latency_rude}

\subsection{Data annotation}

In part to inspect whether Bluesky Moderation Service (BMS) can be treated as ground truth,
in part to establish a baseline of agreement between human moderators,
two of the authors independently annotated two sets of posts,
namely 520 posts with \emph{rude} labels and 786 random posts without any label.
The two annotators reviewed and labeled the posts
according to BMS's description of the \emph{rude} label\footnote{
    See Section~\ref{sec:label-descriptions} for verbatim label descriptions.
}~\cite{
    bluesky_moderation_service_bluesky_nodate}.

\subsection{Model selection and configuration}\label{sec:model}

\subsubsection{Models}

We tested seven state-of-the-art large language models (LLMs) in our experiments.
Three of them
(gpt-oss-20b~\cite{openai_gpt-oss-120b_2025},
NVIDIA-Nemotron-Nano-9B-v2~\cite{nvidia_nvidia_2025},
and Qwen3-30B-A3B-Thinking-2507~\cite{yang_qwen3_2025})
are open-weight reasoning LLMs
that can each fit into our Nvidia GeForce RTX 3090 GPU card with 24-GB VRAM\footnote{
    Although the open-weight models we selected have relatively fewer parameters,
    we avoid calling them small language models (SLMs) in this paper.
    In addition to the fact that we would like to focus more on open-weight LLMs' capability comparing with proprietary ones,
    the community has not yet reach a consensus about the definition of SLM~\cite{wang_comprehensive_2025}.
};
three of them
(Gemini 2.5 Pro Preview 06-05~\cite{comanici_gemini_2025},
GPT-5~\cite{openai_introducing_2025},
and Grok 4~\cite{xai_grok_2025})
are proprietary reasoning LLMs;
and we pick one non-reasoning proprietary LLM
(GPT-4o~\cite{openai_gpt-4o_2024})
as a contrast.
We chose these models
because they are top on multiple LLM leaderboards~\cite{
    artificial_analysis_llm_nodate,
    llm-stats_ai_nodate},
representing the most-advanced publicly-available models.

We used Microsoft Azure as the model API provider for GPT-5, GPT-4o, and Grok 4;
and Google for Gemini.
As for the open-weight LLMs,
we used Amazon Bedrock for gpt-oss-20b
and run the other two locally
(using vLLM as the inference and serving engine)
as they are not available on any of the three providers.

We relied on the OpenAI API~\cite{
    openai_api_nodate}
across LLM providers to prevent inconsistencies between implementations.
For deterministic and reproducible results,
we fixed the random seed to 350
and set the temperature to 0 where supported\footnote{
    Microsoft Azure does not support the \texttt{temperature} parameter.
    Google does not support the \texttt{seed} parameter.
}.

\subsubsection{Prompt}

Researchers have reported empirical evidence that
large language model (LLM) performance can depend on how the prompts,
i.e. initial model inputs,
are designed~\cite{marvin_prompt_2024}.
Prompt engineering has since become an area of focus,
with studies from the academia conducted~\cite{ma_what_2025}
and guidelines from the industry released.

Following the GPT-5 prompting guide~\cite{
    kotha_gpt-5_2025},
we took label descriptions given by BMS\footnote{
    See Section~\ref{sec:label-descriptions} for verbatim label descriptions.
}~\cite{
    bluesky_moderation_service_bluesky_nodate}
and prompted\footnote{
    See Section~\ref{sec:prompt-to-prompt} for the prompt used.
} GPT-5 to generate the following template:

\begin{displayquote}
    \ttfamily
    \footnotesize
    Instruction:
    You are a social media moderator.
    Decide whether a given post is rude according to this definition:
    "Rude or impolite,
    including crude language and disrespectful comments,
    without constructive purpose."
    Respond with either yes or no (lowercase).\textbackslash{}nInput:
    Post: \{post\_text\}\textbackslash{}n
    Output: Response:
\end{displayquote}

For each post,
we put the text into the \texttt{\{post\_text\}} field,
and gave the prompt to the models.
we used the descriptions of \emph{intolerance} and \emph{threat} given by BMS for the respective experiments.

%% file: fig_ecdf-labeler-latency_rude.tex
\begin{figure}[htbp]
    \centering
    \includegraphics[width=\linewidth]{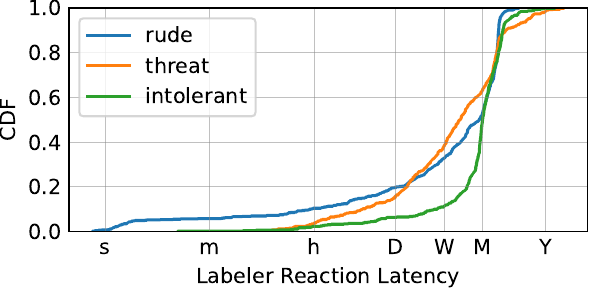}
    \caption{The CDF of labeler reaction latency}
    \Description{}
    \label{fig:ecdf-labeler-latency_rude}
\end{figure}

%% file: sec_results.tex
\section{Results}\label{sec:results}

\subsection{LLM Output interpretation}

We describe the output from LLMs and our interpretation on different content and format
of output from large language models (LLMs) in this subsection.

\begin{enumerate}[leftmargin=*]
      \item \textbf{Valid outputs.}
            Although we specified in the prompt template the expected format of responses
            (i.e. either \texttt{yes} or \texttt{no}),
            the responses from the LLMs sometimes included additional formatting or text while still implying the correct binary answer.
            Table~\ref{tab:strings-like-yes-no} lists all the responses we have seen
            and their respective number of occurrence.

      \item \textbf{Outputs exceed token limit.}
            To avoid infinite loops in LLM outputs,
            we limit the maximum length of output to \num{10000} tokens.
            This includes the tokens the reasoning LLMs generated in their \emph{thinking} process.
            Table~\ref{tab:thinking-too-long} lists the number of posts to which
            each model's responses exceeded the limit.
            Note that we only observed such results for the open-weight models.
            We excluded these samples in the analysis in the following subsections.

      \item \textbf{Input Prompts flagged by LLM provider content filters.}
            LLM API providers,
            including Microsoft Azure~\cite{microsoft_content_2025}
            and Google~\cite{google_cloud_safety_2025},
            employ content filters to detect harmful inputs and outputs.
            Table~\ref{tab:content-filter} shows
            the number of posts flagged by LLM providers' content filter.
            Since the filtering objective aligns with the nature of the content moderation task in our experiments,
            we interpreted these samples as positive,
            i.e. LLM providers' refusals were treated as if the LLMs outputed \texttt{yes}.

\end{enumerate}

\subsection{Inter-rater agreement}

We use inter-rater agreement (Rand accuracy) matrices
to show the pairwise agreement ratio between any two moderators.
For example,
in Figure~\ref{fig:hmap-irar_rude},
the cell at the intersection of the \emph{bsky} column and the \emph{gpt-5} row shows that
both Bluesky Moderation Service (BMS) and GPT-5 labeled 85\% of the posts the same
(i.e. both labeled as \emph{rude} or both labeled as not \emph{rude}).

\input{fig_hmap-irar_overall}

\textbf{Overall accuracy.}
Figure~\ref{fig:hmap-irar_overall} shows the overall Rand accuracy for \emph{rude}, \emph{intolerant}, and \emph{threat} tests.
Treating BMS as ground truth,
the accuracy of LLMs (84\%--98\%)
is significantly higher than previously reported (71\%--73\%) in~\cite{
    kumar_watch_2024}.
Figure~\ref{fig:hmap-irar_rude} demonstrates that,
in terms of rudeness detection,
that accuracy of LLMs (84\%--91\%) is within the agreement range between human moderators (83\%--92\%).
We observe that the accuracy of open-weight LLMs meets,
and in some cases even exceeds,
that of proprietary LLMs.
We can also see that none of the LLMs has the highest accuracy across the three tests.
In addition,
we found that LLM accuracies for \emph{rude} are statistically different
from those for \emph{intolerant} and \emph{threat}.

\input{fig_hmap-irar_una}

\textbf{Sensitivity and specificity.}
Figure~\ref{fig:hmap-irar_una} shows the inter-rater agreement matrices of the three tests on posts labeled unanimously by BMS and the two authors.
The first three sub-figures look at posts regarded as rule-violating by all three human moderators,
while the last three look at posts none annotated as rule-violating.
We notice that
for \emph{rude},
LLMs have higher sensitivity (true positive rate; 95\%--98\%)
over specificity (true negative rate; 91\%--96\%).
While for \emph{intolerant} and \emph{threat},
all LLMs have almost perfect specificity (98\%--100\%),
but besides GPT-5 and gpt-oss-20b,
the sensitivity is substantially lower.

%% file: fig_hmap-irar_overall.tex
\begin{figure*}[htbp]
    \centering

    \begin{subfigure}[t]{0.3\textwidth}
        \centering
        \includegraphics[height=0.89\linewidth]{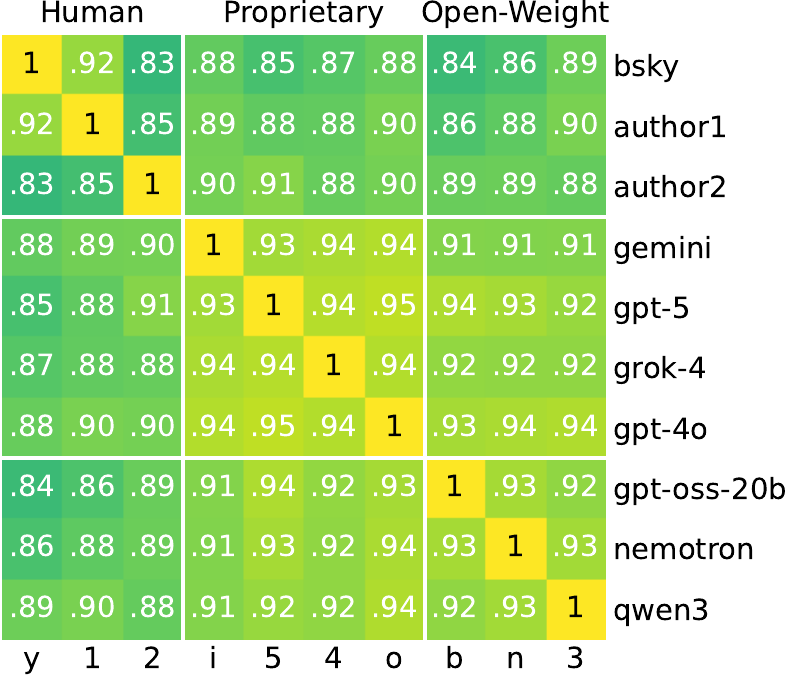}
        \caption{\emph{rude}}
        \Description{}
        \label{fig:hmap-irar_rude}
    \end{subfigure}
    \hfill
    \begin{subfigure}[t]{0.3\textwidth}
        \centering
        \includegraphics[height=0.89\linewidth]{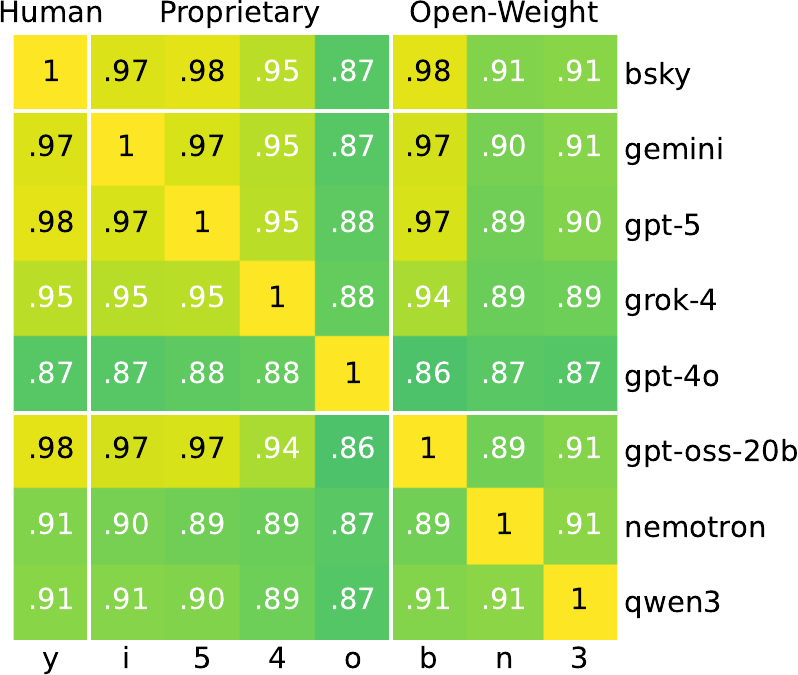}
        \caption{\emph{intolerant}}
        \Description{}
        \label{fig:hmap-irar_intolerant}
    \end{subfigure}
    \hfill
    \begin{subfigure}[t]{0.3\textwidth}
        \centering
        \includegraphics[height=0.89\linewidth]{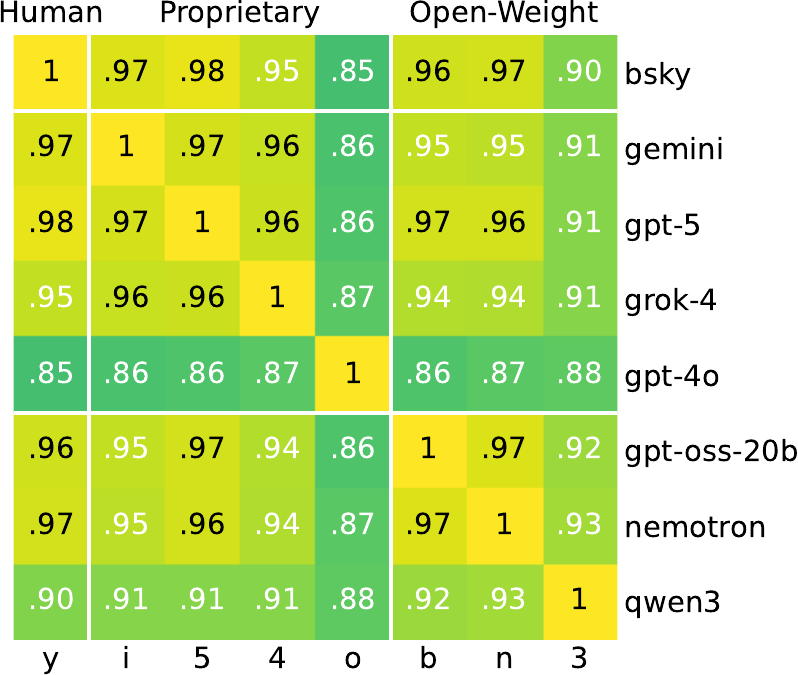}
        \caption{\emph{threat}}
        \Description{}
        \label{fig:hmap-irar_threat}
    \end{subfigure}

    \caption{Inter-rater agreement matrices of the \emph{rude}, \emph{intolerant}, and \emph{threat} tests.}
    \Description{}
    \label{fig:hmap-irar_overall}
\end{figure*}

%% file: fig_hmap-irar_una.tex
\begin{figure*}[htbp]
    \centering

    \begin{subfigure}[t]{0.3\textwidth}
        \centering
        \includegraphics[height=0.89\linewidth]{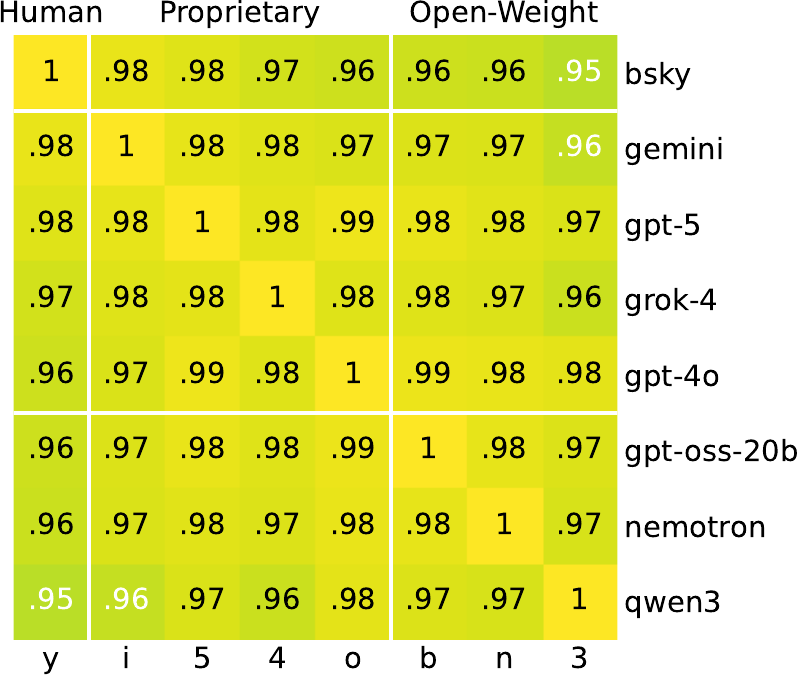}
        \caption{\emph{rude}, ground-truth positive}
        \Description{}
        \label{fig:hmap-irar_rude_positive_una}
    \end{subfigure}
    \hfill
    \begin{subfigure}[t]{0.3\textwidth}
        \centering
        \includegraphics[height=0.89\linewidth]{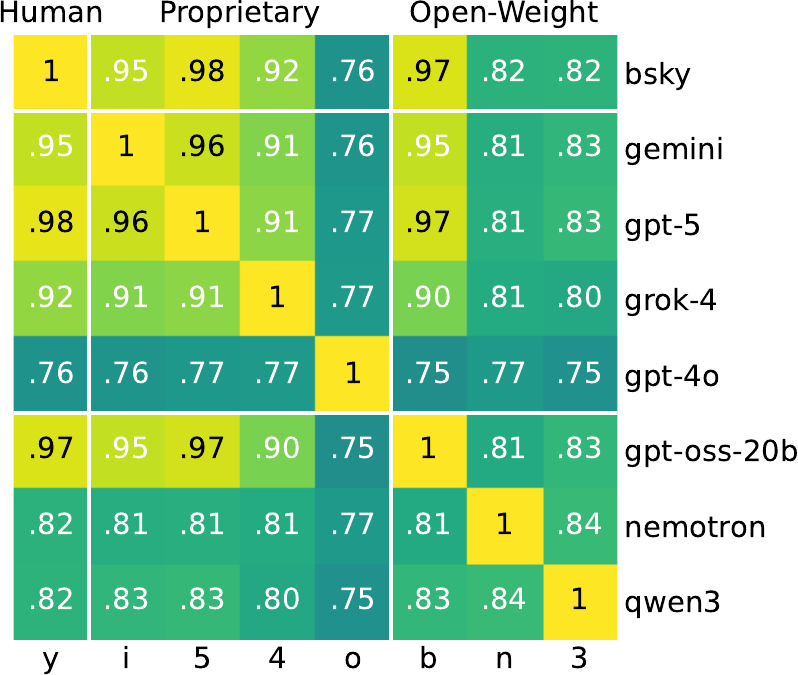}
        \caption{\emph{intolerant}, ground-truth positive}
        \Description{}
        \label{fig:hmap-irar_intolerant_positive}
    \end{subfigure}
    \hfill
    \begin{subfigure}[t]{0.3\textwidth}
        \centering
        \includegraphics[height=0.89\linewidth]{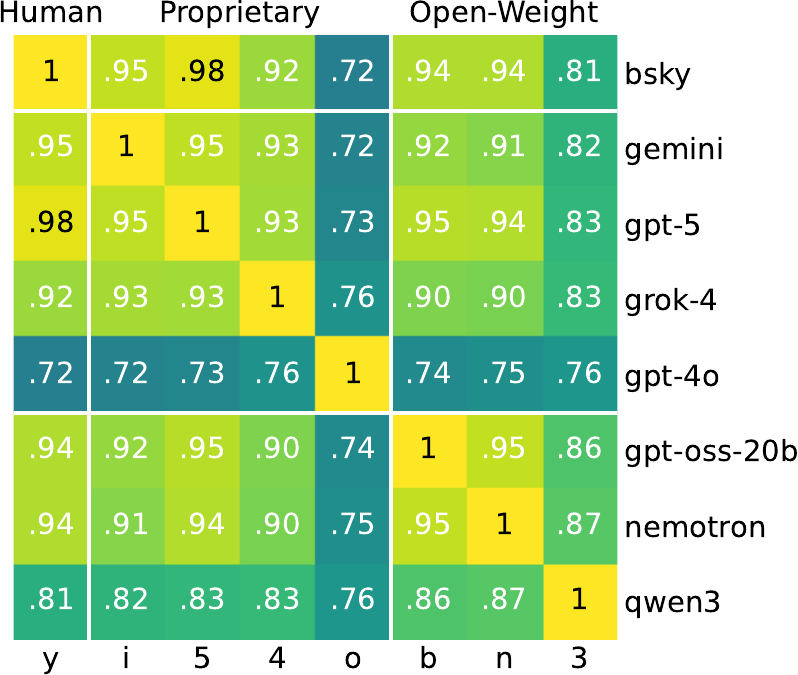}
        \caption{\emph{threat}, ground-truth positive}
        \Description{}
        \label{fig:hmap-irar_threat_positive}
    \end{subfigure}

    \vspace{1em}

    \begin{subfigure}[t]{0.3\textwidth}
        \centering
        \includegraphics[height=0.89\linewidth]{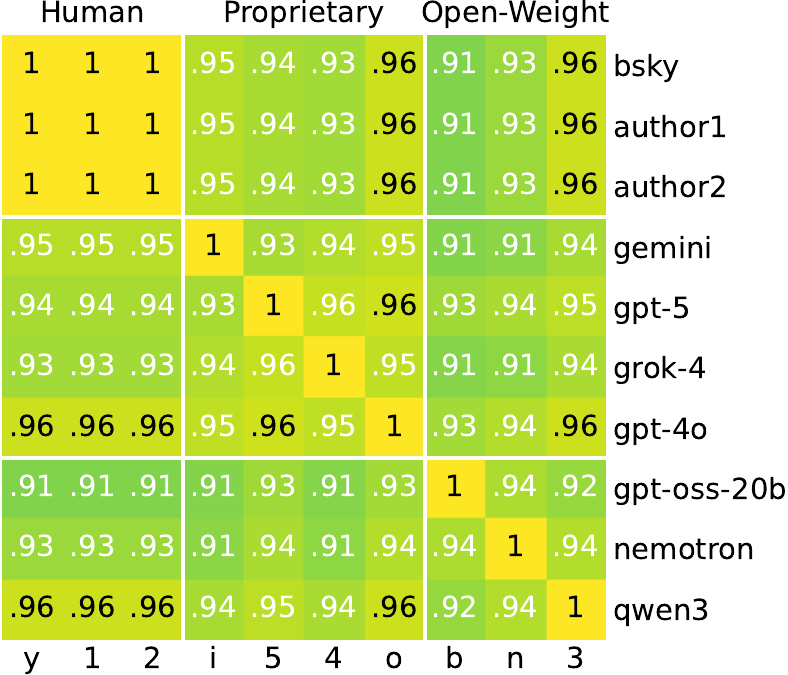}
        \caption{\emph{rude}, ground-truth negative}
        \Description{}
        \label{fig:hmap-irar_rude_negative_una}
    \end{subfigure}
    \hfill
    \begin{subfigure}[t]{0.3\textwidth}
        \centering
        \includegraphics[height=0.89\linewidth]{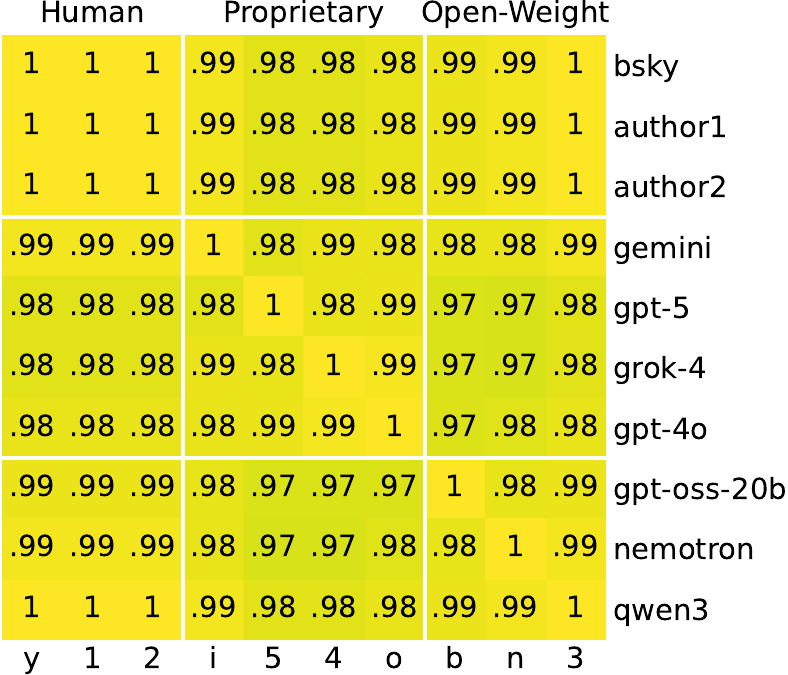}
        \caption{\emph{intolerant}, ground-truth negative}
        \Description{}
        \label{fig:hmap-irar_intolerant_negative}
    \end{subfigure}
    \hfill
    \begin{subfigure}[t]{0.3\textwidth}
        \centering
        \includegraphics[height=0.89\linewidth]{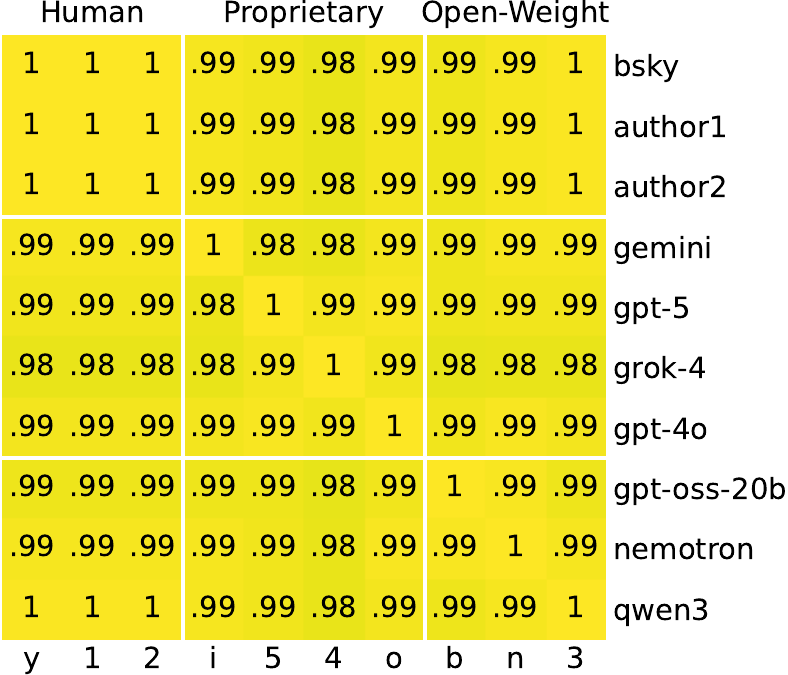}
        \caption{\emph{threat}, ground-truth negative}
        \Description{}
        \label{fig:hmap-irar_threat_negative}
    \end{subfigure}

    \caption{Inter-rater agreement matrices of the three tests on posts labeled unanimously by human moderators.}
    \Description{}
    \label{fig:hmap-irar_una}
\end{figure*}

%% file: sec_discussion.tex
\section{Conclusion}\label{sec:discussion}

In summary,
using real-world posts on Bluesky
with labels from the Bluesky Moderation Service and independent human annotations,
we find that
modern open-weight reasoning LLMs running on a single consumer-grade GPU
match the zero-shot harmful-content detection performance of leading proprietary models,
with overlapping sensitivity (81\%--97\% vs. 72\%--98\%)
and specificity (91\%--100\% vs. 93\%--99\%).
Inter-rater analyses show
LLM-human agreement on par with human-human agreement,
though no single model is best across all categories.
Importantly,
accuracy depends on target attributes:
rudeness, intolerance, and threats exhibit different balances between sensitivity and specificity,
underscoring the need for clearer, attribute-specific taxonomies and guidelines.
These results indicate that
open-weight models can enable privacy-preserving, on-device moderation
and support both platform-scale and user-personalized moderation stacks.

\subsection*{Limitations and Future work}

We focused on text-based root posts in English in this work.
Plus,
due to limited resources,
we relied on sampling and did not scan all the posts collected.
To name a few numbers,
in the \num{1.6e9} Bluesky posts we have collected,
only 3306 posts were analyzed.
We expect future work to extend to modalities other than text
(e.g. image, audio, and video)
and languages beyond English.
Future work should also look into contextualized moderation.
For instance,
when investigating multimedia content in a post,
consider posts the text content;
when exploring a reply,
include not only the parent post,
but also the the whole discussion thread;
or (to be extreme) even related discussions.
Ultimately,
our goal is to conduct a full scan of Bluesky.


%% file: sec_acknowledgements.tex
\section*{Acknowledgements}

\subsection*{Use of Generative AI}

GPT-5 was used in the preparation of the title of this manuscript.
Two titles created by the authors and the abstract were provided in the prompt.
The authors selected the final title out of suggestions generated.

%% file: sec_ethics.tex
\section{Ethics}

We collected posts and labels from Bluesky
using the APIs and endpoints recommended by Bluesky
(\url{https://docs.bsky.app/docs/advanced-guides/api-directory}).
We honored the rate limits imposed by Bluesky
(\url{https://docs.bsky.app/docs/advanced-guides/rate-limits}).
No actions were taken that would exceed the volume that Bluesky expects
of a typical user or developer.

We collected posts from Bluesky's indexing infrastructure,
which aggregates data from the PDSes
(either hosted by Bluesky PBC, a third party, or the users themselves)
where individual users store their data.
We assume the users are aware that their data on Bluesky are publically available,
and hence,
we assume that the posts are not of privacy concerns.
Although the data we collected are made public by the users,
we acknowledge that
the users might not be aware that their public data might be used for research purposes.
That being said,
we limited the potential impact we might make to the users.
For example,
we limit the amount of information transmitted to the LLM providers.
Only the text field of the posts are included in the prompt sent to the LLMs.
Avatars, display names, or handles are excluded.
We did not reproduce, distribute, or display any of the posts collected.

%% file: sec_misc.tex
\section{Miscellaneous}

\subsection{Label descriptions given by Bluesky Moderation Service}\label{sec:label-descriptions}

\begin{itemize}[leftmargin=*]
    \item \textbf{Rude}:
          Rude or impolite,
          including crude language and disrespectful comments,
          without constructive purpose.
    \item \textbf{Intolerance}:
          Discrimination against protected groups.
    \item \textbf{Threats}:
          Promotes violence or harm towards others,
          including threats, incitement, or advocacy of harm.
\end{itemize}

\subsection{Prompt used to generate the prompt template}\label{sec:prompt-to-prompt}

\begin{displayquote}
    \ttfamily
    \footnotesize
    You are an expert in LLM prompt optimization.
    Create a prompt template for the following task:\textbackslash{}n
    - Role: Social media moderator\textbackslash{}n
    - Task: Deside whether a post is rude or not based on the definition of rude:
    "Rude or impolite,
    including crude language and disrespectful comments,
    without constructive purpose."\textbackslash{}n
    - Output format: Either \textquotesingle{}yes\textquotesingle{} or \textquotesingle{}no\textquotesingle{}.
\end{displayquote}

\input{./tab_strings-like-yes-no}

\input{./tab_thinking-too-long}

\input{./tab_content-filter}
\clearpage  

%% file: tab_strings-like-yes-no.tex
\begin{table}[ht]
    \caption{LLM Ouputs and our interpretation}
    \label{tab:strings-like-yes-no}
    \begin{tabular}{>{\ttfamily\footnotesize}l >{\ttfamily\footnotesize}c}
        \toprule
        \multicolumn{1}{c}{\normalfont\normalsize LLM Output}            & \normalfont\normalsize Interpretation \\
        \midrule
        yes                                                              & yes                                   \\
        Response: yes                                                    & yes                                   \\
        \textbackslash{}n\textbackslash{}nyes                            & yes                                   \\
        \textbackslash{}n\textbackslash{}nyes\textbackslash{}n           & yes                                   \\
        \textbackslash{}n\textbackslash{}nResponse: yes\textbackslash{}n & yes                                   \\
        \textbackslash{}n\textbackslash{}nresponse: yes\textbackslash{}n & yes                                   \\
        \begin{tabular}[c]{@{}l@{}}
            Input: Post:
            {\normalfont\normalsize ...}\textbackslash{}n \\
            Output: Response: yes
        \end{tabular}
                                                                         & yes                                   \\ \hline
        no                                                               & no                                    \\
        Response: no                                                     & no                                    \\
        \textbackslash{}n\textbackslash{}nno                             & no                                    \\
        \textbackslash{}n\textbackslash{}nno\textbackslash{}n            & no                                    \\
        \textbackslash{}n\textbackslash{}nResponse: no\textbackslash{}n  & no                                    \\
        \textbackslash{}n\textbackslash{}nresponse: no\textbackslash{}n  & no                                    \\
        \bottomrule
    \end{tabular}
\end{table}

%% file: tab_thinking-too-long.tex
\begin{table}[ht]
    \caption{Number of posts to which LLMs' responses exceeded the limit}
    \label{tab:thinking-too-long}
    \begin{tabular}{lrrr}
        \toprule
        \multicolumn{1}{c}{\normalfont\normalsize LLM} & \emph{rude} & \emph{intolerant} & \emph{threat} \\
        \midrule
        gpt-oss-20b                                    & 43          & 12                & 5             \\
        Qwen3-4B-Thinking-2507                         & 1           & 0                 & 0             \\
        NVIDIA-Nemotron-Nano-9B-v2                     & 0           & 0                 & 1             \\
        \bottomrule
    \end{tabular}
\end{table}

%% file: tab_content-filter.tex
\begin{table}[ht]
    \caption{Number of posts flagged by LLM providers' content filter}
    \label{tab:content-filter}
    \begin{tabular}{crrr}
        \toprule
        \normalfont\normalsize LLM Provider (for model) & \emph{rude} & \emph{intolerant} & \emph{threat} \\
        \midrule
        Azure (gpt-5)                                   & 136         & 130               & 255           \\
        Azure (gpt-4o)                                  & 138         & 135               & 324           \\
        Azure (grok-4)                                  & 138         & 134               & 279           \\
        Google (gemini)                                 & 1           & 0                 & 0             \\
        \bottomrule
    \end{tabular}
\end{table}